\documentclass{llncs}

\usepackage{tabu}
\usepackage[utf8]{inputenc}

\usepackage{algpseudocode}
\usepackage{graphicx}
\usepackage{multirow}
\usepackage{enumitem}
\usepackage{adjustbox}

\usepackage{algorithm}

\usepackage{url}

\usepackage{times}

\usepackage{latexsym}
\usepackage{comment}
\usepackage{csquotes}

\usepackage{epsfig}
\usepackage[colorinlistoftodos]{todonotes}
\usepackage{graphicx}
\usepackage{amsmath}
\usepackage{float}

\begin{document}
\mainmatter

\title{Composite Semantic Relation Classification}

\author{Siamak Barzegar\inst{1} \and Andre Freitas\inst{2} \and Siegfried Handschuh\inst{2} \and Brian Davis\inst{1}}

\institute{Insight Centre for Data Analytics \\
	National University of Ireland, Galway \\ Galway, Ireland \\
	\texttt{\{Firstname.Lastname\}@insight-centre.org} \\
	\and
	Digital Libraries and Web Information Systems \\
	University of Passau
    \\ Passau, Germany \\
	\texttt{\{Fistname.Lastname\}@uni-passau.de} \\}

\maketitle

\begin{abstract}

Different semantic interpretation tasks such as text entailment and question answering require the classification of semantic relations between terms or entities within text. However, in most cases it is not possible to assign a direct semantic relation between entities/terms. This paper proposes an approach for composite semantic relation classification, extending the traditional semantic relation classification task. Different from existing approaches, which use machine learning models built over lexical and distributional word vector features, the proposed model uses the combination of a large commonsense knowledge base of binary relations, a distributional navigational algorithm and sequence classification to provide a solution for the composite semantic relation classification problem.

% 	Although distributional semantic models can effectively capture the semantic relatedness between two entities/terms, they are limited with respect to capturing the type of relation.  Further more an explicit direct relation between entities does not always exists. The ability to compute both relatedness and relation type aspires to human-like performance to achieve the goal of producing meaningful representations of text.  Therefore, to cope with this challenge, the goal is to learn composition semantic relations. 
% One of the methodologies is used is bring a systematic way to evaluate Distributional Semantic models under relation classification. The resources for evaluating them and methodology are not fully available at the moment. In this work, besides proposing a new approach for composite semantic relation classification, a unique dataset is created manually.

% * <brian.davis@insight-centre.org> 2016-12-14T21:56:58.676Z:
%
% >  Therefore, to cope with this challenge, the goal is to learn composition semantic relations. 
%
% This is too short.  Tell me more about what a composition relations is what you did and briefly what the results show.
%
% ^.
\end{abstract}

\section{Introduction}
Capturing the semantic relationship between two concepts is a fundamental operation for many semantic interpretation tasks. This is a task which humans perform rapidly and reliably by using their linguistic and commonsense knowledge about entities and relations. Natural language processing systems which aspire to reach the goal of producing meaningful representations of text must be equipped to identify and learn semantic relations in the documents they process.

The automatic recognition of semantic relations has many applications such as information extraction, document summarization, machine translation, or the construction of thesauri and semantic networks. It can also facilitate auxiliary tasks such as word sense disambiguation, language modeling, paraphrasing, and recognizing textual entailment \cite{hendrickx2009semeval}.

However it is not always possible to establish a direct semantic relation given two entity mentions in text. In the Semeval 2010 Task 8 test collection \cite{hendrickx2009semeval} for example 17.39\% of the semantic relations mapped within sentences were assigned with the label \textit{"OTHER"}, meaning that they could not be mapped to the set of 9 direct semantic relations \footnote{Cause-Effect,  Instrument-Agency,  Product-Producer, Content-Container, Entity-Origin, Entity-Destination,  Component-Whole,  Member-Collection,  Communication-Topic}. 
In many cases, the semantic relations between two entities can only be expressed by a composition of two or more operations. This work aims at improving the description and the formalization of the semantic relation classification task by introducing the concept of composite semantic relation classification, in which the relations between entities can be expressed using the composition of one or more relations. 

This paper is organized as follows: Section \ref{csrc} describes the semantic relation classification problem and the related work followed by the proposed composite semantic relation classification (Section \ref{proposed}), Section \ref{baseline} describes the
existing baseline models; while Section \ref{evaluation} describes the experimental setup and analyses the results, providing a comparative analysis between the proposed model and the baselines. Finally, Section \ref{Conclusion} provides the conclusion.

\section{Composite Semantic Relation Classification}\label{csrc}

\subsection{Semantic Relation Classification} 

Semantic relation classification is the task of classifying the underlying abstract semantic relations between target entities (terms) present in texts \cite{qin2016empirical}. The goal of relation classification is defined as follows: given a sentence $S$ with the pairs of annotated target nominals $e_1$ and $e_2$, the relation classification system aims to classify the relations between $e_1$ and $e_2$ in given texts within the pre-defined relation set \cite{hendrickx2009semeval}. For instance, the relation between the nominal \textbf{burst} and \textbf{pressure} in the following example sentence is interpreted as \textbf{Cause-Effect($e_2, e_1$)}.

\begin{displayquote}
	The $<e_1>burst</e_1>$ has been caused by water hammer $<e_2>pressure</e_2>$.
\end{displayquote}

\subsection{Existing Approaches for Semantic Relation Classification}

Different approaches have been explored for relation classification, including unsupervised relation discovery and supervised classification. Existing literature have proposed various features to identify the relations between entities using different methods.

Recently, {Neural network-based approaches}  have achieved significant improvement over traditional methods based on either human-designed features\cite{qin2016empirical}. However, existing neural networks for relation classification are usually based on shallow architectures (e.g., one-layer convolutional neural networks or recurrent networks). In exploring the potential representation space at different abstraction levels, they may fail to perform\cite{xu2016improved}. 

The performance of supervised approaches strongly depends on the quality of the designed features \cite{zeng2014relation}. With the recent improvement in Deep Neural Network (DNN), many researchers are experimenting with unsupervised methods for automatic feature learning. \cite{xu2015classifying} introduce gated recurrent networks, in particular, Long short-term memory (LSTM), to relation classification. \cite{zeng2014relation} use Convolutional Neural Network (CNNs). Additionally, \cite{dos2015classifying} replace the common Softmax loss function with a ranking loss in their CNN model. \cite{xu2015semantic} design a negative sampling method based on CNNs. From the viewpoint of model ensembling, \cite{liu2015dependency} combine CNNs and recursive networks along the Shortest Dependency Path (SDP), while \cite{nguyen2015combining} incorporate CNNs with Recurrent Neural Networks (RNNs). 

Additionally, much effort has been invested in relational learning methods that can scale to large knowledge bases. The best performing neural-embedding models are Socher(NTN)\cite{socher2013reasoning} and Bordes models (TransE and TATEC) \cite{bordes2013translating,garcia2016combining}.

\section{From Single to Composite Relation Classification} \label{proposed}

\subsection{Introduction}

The goal of this work is to propose an approach for semantic relation classification using one or more relations between term mentions/entities.

\begin{displayquote}
	"The $<e_1>child</e_1>$ was carefully wrapped and bound into the  $<e_2>cradle</e_2>$ by means of a cord."
\end{displayquote}

In this example, the relationship between $Child$ and $Cradle$ cannot be directly expressed by one of the nine abstract semantic relations from the set described in \cite{hendrickx2009semeval}. %Remove Table and cite to Paper NLDB
%[!htbp]

However, looking into a commonsense KB (in this case, ConceptNet V5.4) we can see the following set of composite relations between these elements:
% nldb
\begin{displayquote}
	$<e_1>child</e_1>$ $createdby \circ causes \circ atlocation$ $<e_2>cradle</e_2>$
\end{displayquote}

As you increase the number of edges that you can include in the set of semantic relations compositions (the size of the semantic relationship path), there is a dramatic increase in the number of paths which connect the two entities. For example, for the words $Child$ and $Cradle$ there are 15 paths of size 2, 1079 paths of size 3 and 95380 paths of size 4. Additionally, as the path size grows many non-relevant relationships (less meaningful relations) will be included.

The challenge in \textit{composite semantic relation classification} is to provide a classification method that provides the most meaningful set of relations for the context at hand. This task can be challenging because, as previously mentioned, a simple KB lookup based approach would provide all semantic associations at hand.

To achieve this goal we propose an approach which combines \textit{sequence machine learning models}, \textit{distributional semantic models} and \textit{commonsense relations  knowledge bases} to provide an accurate method for composite semantic relation classification.

\begin{figure}[ht]
\centering
\includegraphics[height=6.2cm]{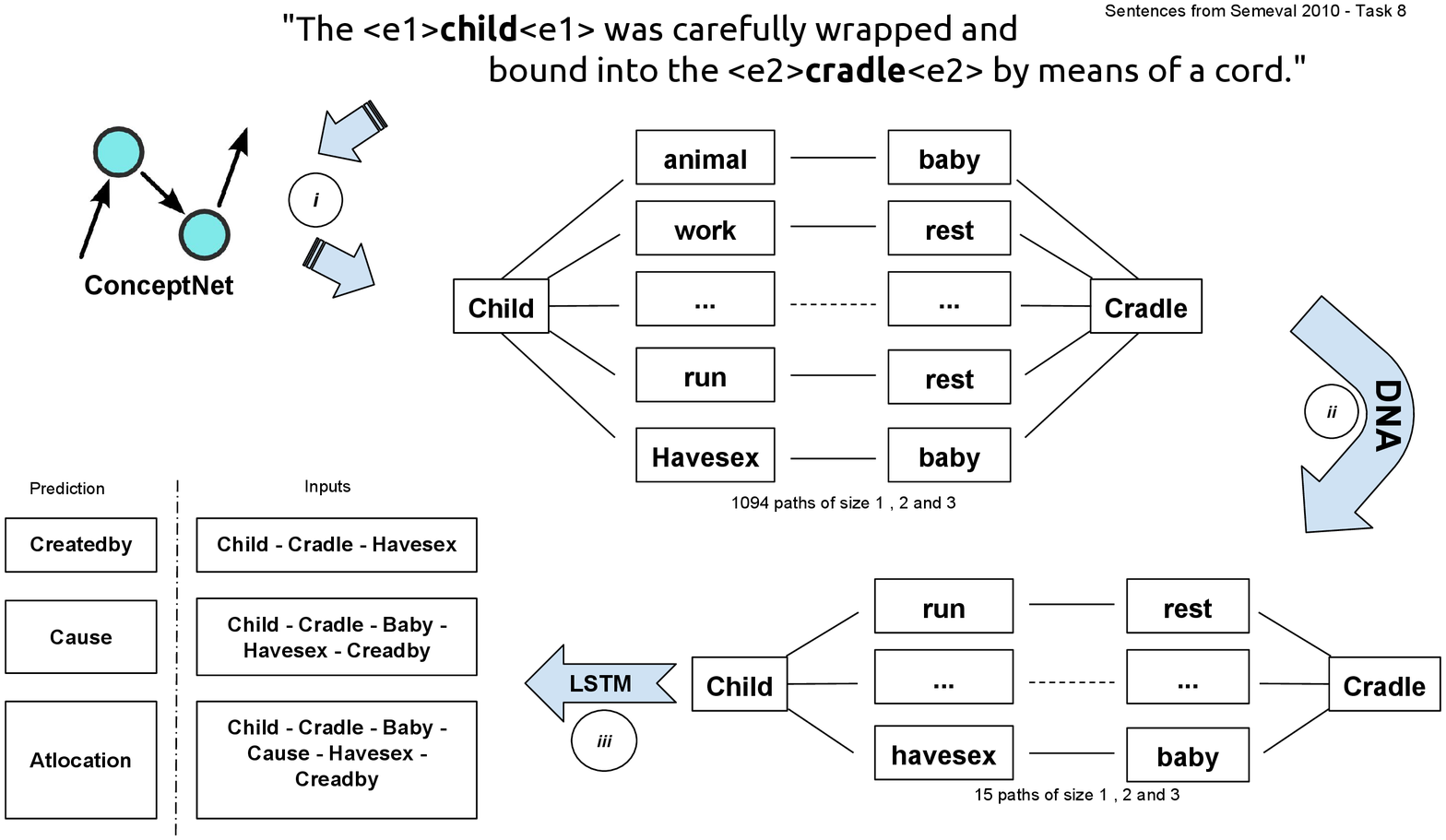}
\caption{Depiction of the proposed model relies on the combination of the our three approaches}
\label{fig:proposed_model}
\end{figure}

The proposed model (Fig \ref{fig:proposed_model}) relies on the combination of the following approaches:

\begin{enumerate}[label=\roman*]
\item Use existing structured commonsense KBs define an initial set of semantic relation compositions.
\item Use a pre-filtering method based on the Distributional Navigational Algorithm (DNA) as proposed by \cite{freitas2014distributional}
\item Use sequence-based Neural Network based model to quantify the sequence probabilities of the semantic relation compositions. We call this model Neural Concept/Relation Model, in analogy to a Language Model.

\end{enumerate}
%[!htbp]

\subsection{Commonsense KB Lookup}\label{commensense_KB}

The first step consists in the use of a large commonsense knowledge base for providing a reference for a sequence of semantic relations. ConceptNet is a semantic network built from existing linguistic resources and crowd-sourced. It is built from nodes representing words or short phrases of natural language, and labeled abstract relationships between them.

1094 paths were extracted from ConceptNet with two given entities (e.g. $child$ and $cradle$) with no corresponding semantic relation from the Semeval 2010 Task 8 test collection (Figure \ref{fig:proposed_model}(i)). Examples of paths are: 

\begin{itemize}
\item \textbf{child/canbe/baby/atlocation/cradle}
\item child/isa/animal/hasa/baby/atlocation/cradle
\item child/hasproperty/work/causesdesire/rest/synonym/cradle
\item child/instanceof/person/desires/baby/atlocation/cradle
\item child/desireof/run/causesdesire/rest/synonym/cradle
\item \textbf{child/createdby/havesex/causes/baby/atlocation/cradle}
\end{itemize}

\subsection{Distributional Navigational Algorithm (DNA)}

The Distributional Navigational Algorithm (DNA) consists of an approach which uses distributional semantic models as a relevance-based heuristic for selecting relevant facts attached to a contextual query. The approach focuses on addressing the following problems: (i) providing a semantic selection mechanism for facts which are relevant and meaningful in a particular reasoning \& querying context and (ii) allowing coping with information incompleteness in a huge KBs.

In \cite{freitas2014distributional} DSMs are used as a complementary semantic layer to the relational model, which supports coping with semantic approximation and incompleteness.

For large-scale and open domain commonsense reasoning scenarios, model completeness, and full materialization cannot be assumed. A commonsense KB would contain vast amounts of facts, and a complete inference over the entire KB would not scale to its size. Although several meaningful paths may exist between two entities,  there are a large number of paths which are not meaningful in a specific context. For instance, the reasoning path which goes through (1) is not related to the goal of the entity pairs (the relation between $Child$ of human and $Cradle$) and should be eliminated by the application of the Distributional Navigation Algorithm (DNA) \cite{freitas2014distributional}, which computes the distributional semantic relatedness between the entities and the intermediate entities in the KB path as a measure of semantic coherence. In this case the algorithm navigates from $e1$ in the direction of $e2$ in the KB using distributional semantic relatedness between the target node $e2$ and the intermediate nodes $en$ as a heuristic method.
%[!htbp]
\begin{figure}[ht]
\centering
\includegraphics[height=5.2cm]{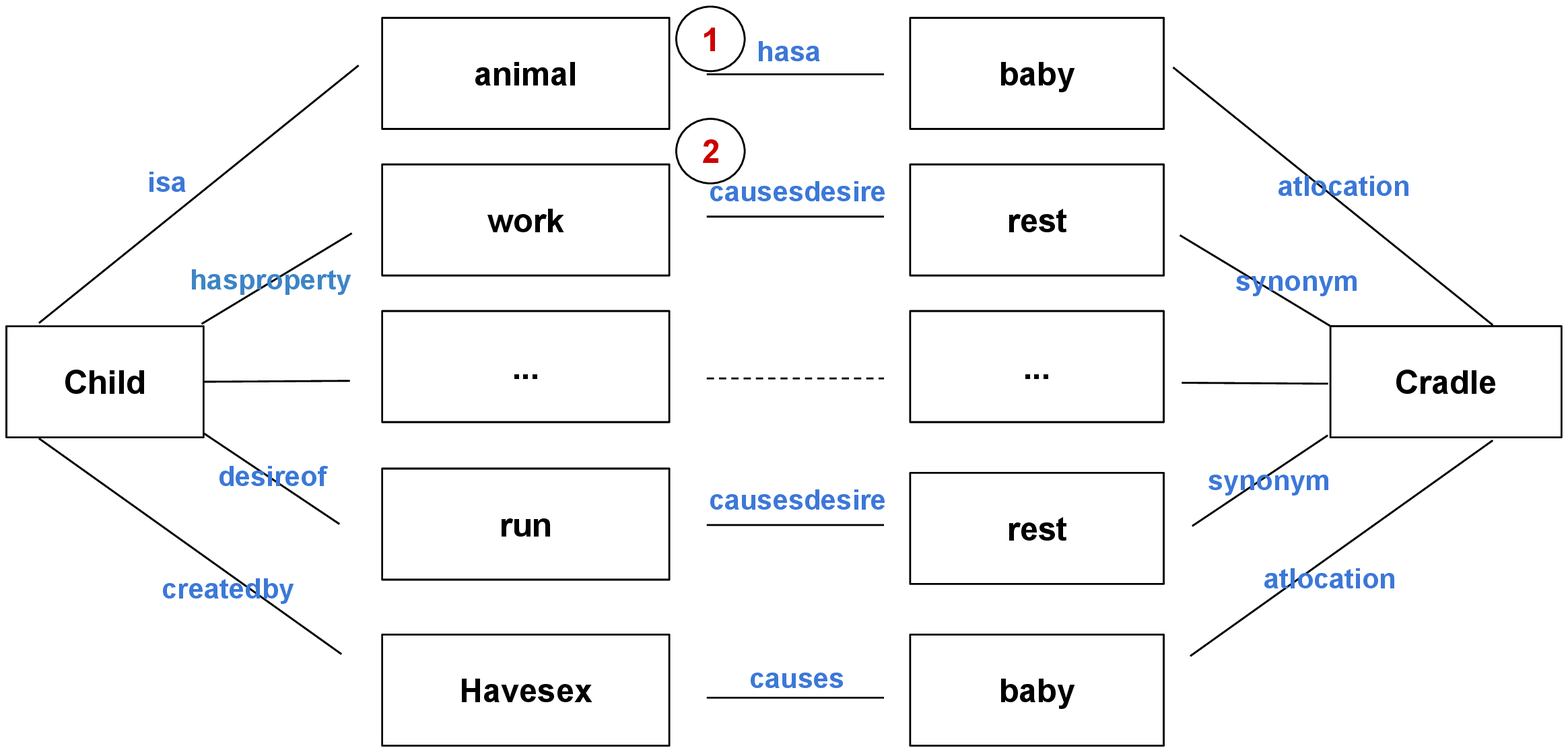}
\caption{Selection of meaningful paths}
\label{exampleKB}
\end{figure}

\subsection{Neural Entity/Relation Model}

\begin{figure}[ht]
\centering
\includegraphics[height=12.2cm]{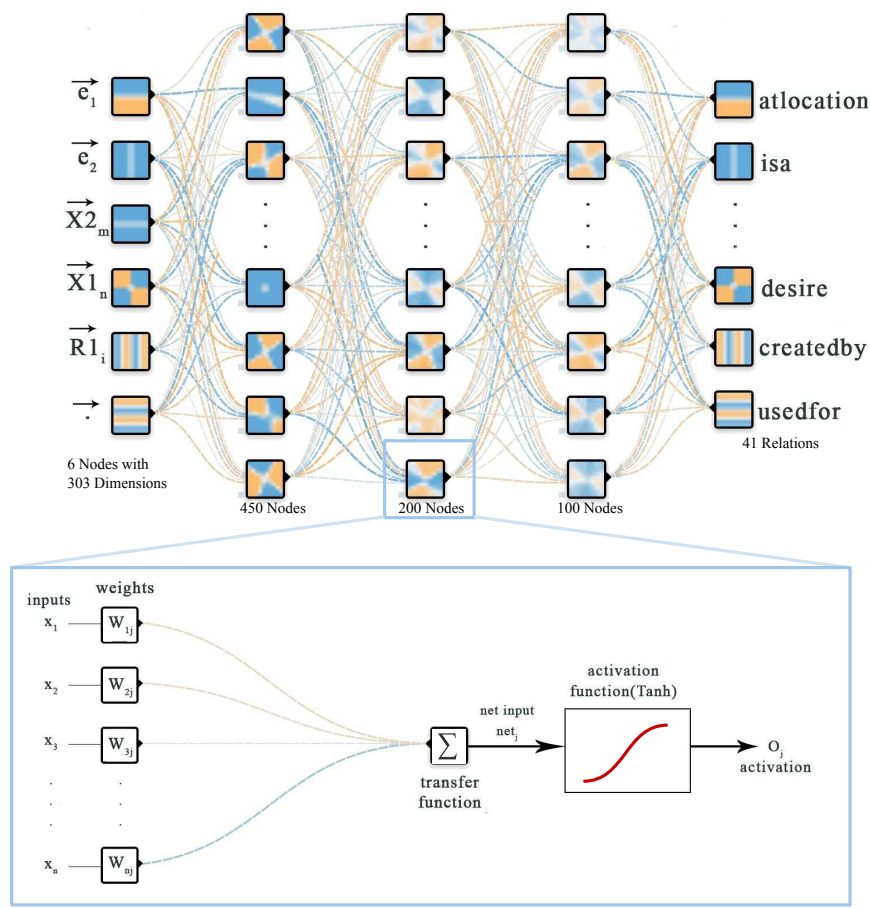}
\caption{The LSTM-CSRC architecture}
\label{fig:LSTM}
\end{figure}

The Distributional Navigational Algorithm provides a pre-filtering of the relations maximizing the semantic relatedness coherence. This can be complemented by a predictive model which takes into account the likelihood of a sequence of relations, i.e. the likelihood of a composition sequence. The goal is to systematically compute the sequence of probabilities of a relation composition, in a similar fashion to a language model. For this purpose we use a Long short-term memory (LSTM) recurrent neural network architecture (Figure \ref{fig:LSTM}) \cite{hochreiter1997long}.

%[ht]
\begin{algorithm}[ht]
	\caption{Composite Semantic Relation Classification}
    \label{euclid}
	\begin{algorithmic}
		\State $I : $ \textit{sentences of semeval 2010-Task 8 dataset}
		\State $O : $ \textit{predefined entity pairs ($e_1$, $e_2$)}
		\State $W :$ \textit{words in I}
		\State $R :$ \textit{related relations of $w$}
		
		\ForAll{$s \in I$}:
		\State $S \gets $ \textit{If entities of} $s$ \textit{are connected in a } $OTHER$ \textit{ relation}
		\EndFor
		
		\ForAll{$s \in S$}:
		\State $ ep \gets$ \textit{predefined entity pairs of }$ s $
		\State $ p \gets $ \textit{find all path of} $ ep $ \textit{in ConceptNet (with maximum paths of size 3)}
		\ForAll{$i \in p$}:
		\State $sq_i  \gets $ \textit{avg similarity score between each word pairs \cite{barzegar2015dinfra} }
		\EndFor
		\State $ msq \gets$ \textit{find max } $sq$
		
		\ForAll{$i \in p$}:
		\State  \textit{filter } $i$ \textit{ If } $sq_i <$ $msq$ - $\frac{msq}{2}$
		\EndFor
		\State $ dw \gets $ \textit{convert } $s$ \textit{ into suitable format for deep learning}
		\EndFor
		
		\State $ model \gets$ \textit{learning LSTM with } $dw$ \textit{ dataset}
		
	%	\ForAll{$s \in O$}:
	%	\State $ current \gets$ \textit{ find next related word of} $s$ \textit{ in semantic network}
	%	\State $ Relation_c  \gets $ (($e_1$, $e_2$, current, previous $W$, previous $R$) $\gets$ model) 
	%	\EndFor
	\end{algorithmic}
\end{algorithm}

\section{Baseline Models}\label{baseline}
As baselines we use bigram language models which define the conditional probabilities between a sequence of semantic relations $r$ after entities $e$, i.e. $P(r \mid e)$.

The performance of baselines systems is measured using the \textit{CSRC\footnote{Composite Semantic Relation Classification} $Cloze$ task}, as defined in section \ref{cloze} where we hold out the last relation and rate a system by its ability to infer this relation.

\begin{itemize}
\item \textbf{Random  Model: } This is the simplest baseline, which outputs randomly selected relation pairs.
\item \textbf{Unigram Model: } Predicts the next relation based on unigram probability of each relation which was calculated from the training set. In this model, relations are assumed to occur independently.
\item \textbf{Single Model: }

The single model is defined by \cite{jans2012skip}:
\begin{equation} \label{eq1}
	\begin{split}
		P(r\mid e) = \frac{P(r, e)}{P(e)}
	\end{split}
\end{equation}
where $P(r \mid e)$ is the probability of seeing $e$ and $r$, in order. 
Let $A$ be an ordered list of relations and entities, $\left | A \right |$ is the length of R, For $i = 1, .., \left | A \right |$, define $a_i$ to be the $ith$ element of A. We rank candidate relations r by maximizing F(r,a), defined as
\begin{equation} \label{eq2}
	\begin{split}
		F(r,a)= \sum_{i=1}^{\left | A \right |-1} log P(r \mid a_i) \end{split}
\end{equation}
where the conditional probabilities $P(r \mid a_i)$ calculated using (1).

\item \textbf{Random Forest: } is an ensemble learning method for classification and other tasks, that operate by constructing a multitude of decision trees at training time and outputting the class that is the mode of the classes. Random decision forests correct for decision trees' habit of overfitting to their training set.
\end{itemize}
\section{Experimental Evaluation}\label{evaluation}
\subsection{Training and Test Dataset} \label{cloze}
The evaluation dataset was generated by collecting all pairs of entity mentions in the Semeval 2010 task 8 \cite{hendrickx2009semeval} which had no attached semantic relation classification (i.e. which contained the relation label \textit{"OTHER"}). 

For all entities with unassigned relation labels, we did a $Conceptnet$ lookup \cite{speer2012representing}, where we generated all paths from sizes 1, 2 and 3 (number of relations) occurring between both entities($e_1$ and $e_2$) and their relations ($R$).

For example:\\
$\textbf{e1} - R1_i - \textbf{e2}$ \\
$\textbf{e1} - R1_i - \textbf{X1}_n - R2_j - \textbf{e2}$ \\
$\textbf{e1} - R1_i - \textbf{X1}_n -R2_j - \textbf{X2}_m - R3_k - \textbf{e2}$  

where $X$ contains the intermediate entities between the target entity mentions \textbf{e1} and \textbf{e2}.

In next step, the Distributional Navigational Algorithm (DNA) is applied over the entity paths\cite{freitas2014distributional}. In the final step of generating training \& test datasets, the best paths are selected manually out of filtered path sets.

From 602 entity pairs assigned to the  \textit{"OTHER"} relation label in Semeval, we found $27,415$ paths between $405$ entity pairs in ConceptNet. With the Distributional Navigation Algorithm (DNA), meaningless paths were eliminated, and after filtering, we have $2,514$ paths for $405$ entity-pairs.

Overall we have $41$ relations and $964$ entities. All paths were converted into the following format which will be input into the neural network: $\textbf{e}_1 - R1_i - \textbf{X1}_n -R2_j - \textbf{X2}_m - R3_k - \textbf{e}_2 $ (Table \ref{Training1}). 

\begin{table}[ht]
    		\caption{Training data-set for CSRC model}
            \label{Training1}
	\begin{center}
		
		\begin{tabular}{|l|l|}
			\hline
			input & Classification \\ \hline
			$\textbf{e}_1$ $\textbf{e}_2$  $\textbf{X1}_n $ & $\textcolor{blue}{\textbf{R1}_i}$ \\ \hline
			$\textbf{e}_1$ $\textbf{e}_2$   $\textbf{X2}_m$ $\textbf{X1}_n$ $\textcolor{blue}{\textbf{R1}_i}$ & $\textcolor{blue}{\textbf{R2}_i}$ \\ \hline
			$\textbf{e}_1$ $\textbf{e}_2$  $\textbf{X2}_m$ $\textcolor{blue}{\textbf{R2}_i}$ $\textbf{X1}_n$ $\textcolor{blue}{\textbf{R1}_i} $ & $\textcolor{blue}{\textbf{R3}_i}$  \\ \hline
		\end{tabular}
	\end{center}
%\hfill
\end{table}

We provide statistics for the generated datasets in the Tables \ref{test-baseline} and \ref{Trainingdataset}. In Table \ref{Trainingdataset} our dataset is divided into a training set and a test set with scale ($75-25\%$), also we used $25$ percent of the training set for cross-validation, $3120$ examples for training, $551$ for validation and $1124$ for testing. Table \ref{test-baseline} shows statistics for test dataset of baseline models.

\begin{table}[ht]
	\caption{Number of different length in the test dataset for baseline models}
    \label{test-baseline} 
	\begin{center}
			\begin{tabular}{|l|l|l|l|}
				\hline
				Test Dataset & \# Length 2 & \# Length 4 & \# Length 6 \\ \hline
				Baselines   & 245 & 391 & 432 \\ \hline
			\end{tabular}
		\end{center}
	\end{table}
	
	\begin{table}[ht]
		\caption{Dataset for LSTM model}
        \label{Trainingdataset}
		\begin{center}
			\begin{tabular}{|l|l|l|l|}
				\hline
				Dataset & \# Train & \# Dev & \# Test \\ \hline
				CSRC         & 3120     & 551    & 1124    \\ \hline
			\end{tabular}
		\end{center}
	\end{table}

	\subsection{Results}

To achieve the classification goal, we generated a LTSM model for the composite relation classification task. In our experiments, a batch size 25, and epoch 50 was generated. An embedding layer using Word2Vec pre-trained vectors was used.

	In our experiment, we optimized the hyperparameters of the LSTM model. After several experiments, the best model is generated with:
    
	\begin{itemize}
		\item Inputs length and dimension are $6$ and $303$, respectively.
		\item Three hidden layers with $450$, $200$ and $100$ nodes and $Tanh$ activation, 
		\item Dropout technique ($0.5$),
		\item $Adam$ optimizer.
	\end{itemize}
	
We experimented our LSTM model with three different pre-training embedding word vector models:
	
	\begin{itemize}
		\item Word2Vec (Google News) with 300 dimensions
		\item Word2Vec (Wikipedia 2016) with 30 dimensions 
		\item No pre-training word embedding
	\end{itemize}
	
The accuracy for the configuration above after 50 epochs is shown in the table below. 
	
	\begin{table}[ht]
		\caption{Validation Accuracy}
        \label{Accuracy}
		\begin{center}
			\begin{tabular}{|l|l|l|l|}
					\hline
					CRSC & W2V Google\_News & W2V Wikipedia & No Pre Training \\ \hline
					Accuracy & 0.4208 &   0.3841  & 0.2196          \\ \hline
				\end{tabular}
			\end{center}
		\end{table}
        
Table \ref{evaluation_results} contains the Precision, Recall, F1-Score and Accuracy.
\begin{table}[!htbp]
\caption{\label{evaluation_results}Evaluation results on baseline models and our approach, with four metrics}
\begin{center}
\begin{tabular}{|l|l|l|l|l|}
\hline
Method         & Recall      & Precision   & F1 Score  & Accuracy    \\ \hline
							Random         & 0.0160 & 0.0220 & 0.0144 & 0.0234 \\ \hline
							Unigram        & 0.0270 & 0.0043 & 0.0074  & 0.1606 \\ \hline
							Single & 0.2613& 0.2944 & 0.2502 & 0.3793 \\ \hline
							Random Forest           & 0.2476 & \textbf{0.3663} & 0.2766 & 0.3299 \\ \hline
							
                            \textbf{LSTM-CSRC}           & \textbf{0.3073} & 0.3281 & \textbf{0.3119} & \textbf{0.4208} \\ \hline
\end{tabular}
\end{center}
\end{table}
Between the evaluated models, the LSTM-CSRC achieved the highest F1 Score and Accuracy. The Single model achieved the second highest accuracy $0.3793$ followed by Random forest model $0.3299$. The LSTM approach provides an improvement of 9.86 \% on accuracy over the baselines, and 11.31 \% improvement on the F1-score. Random Forest achieved the highest precision, while LSTM-CSRC achieved the highest recall.

The extracted information from confusion matrix show in Tables \ref{ConfusionPart1} and \ref{ConfusionPart2}. 
%[!htbp]
\begin{table}[ht]
\caption{The extracted information from Confusion Matrix - Part 1}
\label{ConfusionPart1} 
					\begin{center}
						\resizebox{\textwidth}{!}{
							\begin{tabular}{|l|l|l||l|l|l|}
								\hline
								Relation           & \shortstack{ \# Correct \\ Predicted} & \shortstack{\# Correct \\Predicted Rate} & Relation                     & \shortstack{ \# Correct \\ Predicted}& \shortstack{\# Correct \\Predicted Rate} \\ \hline
								notisa          & 2                    & 1                         & memberof                  & 1                    & 0.5                       \\ \hline
								atlocation      & 172                  & 0.67                      & hasa                      & 24                   & 0.393                     \\ \hline
								notdesires       & 6                    & 0.666                     & hassubevent               & 12                   & 0.378                     \\ \hline
								similar         & 5                    & 0.625                     & partof                    & 16                   & 0.374                     \\ \hline
								desires          & 36                   & 0.593                     & haspropertry              & 12                   & 0.375                     \\ \hline
								hasprerequest   & 23                   & 0.547                     & sysnonym                  & 54                   & 0.312                     \\ \hline
								causesdesire      & 17                   & 0.548                     & derivedfrom               & 20                   & 0.307                     \\ \hline
								isa             & 147                  & 0.492                     & etymologicallyderivedfrom & 6                    & 0.3                       \\ \hline
 								antonym        & 68                   & 0.492                     & capableof                 & 13                   & 0.26                      \\ \hline
								instandof       & 46                   & 0.479                     & motivationbygoal          & 3                    & 0.25                      \\ \hline
 								usedfor         & 47                   & 0.475                     & receivsection             & 5                    & 0.238                     \\ \hline
								desireof        & 5                    & 0.5                       & createdby                 & 4                    & 0.2                       \\ \hline
								hascontext      & 2                    & 0.5                       & madeof                    & 3                    & 0.16                      \\ \hline
								haslastsubevent & 2                    & 0.5                       & causes                    & 3                    & 0.15                      \\ \hline
								nothasa         & 1                    & 0.5                       & genre                     & 1                    & 0.11                      \\ \hline
\end{tabular}}
\end{center}
\end{table}

\begin{table}[ht]
			\caption{The extracted information from Confusion Matrix - Part 2}
            \label{ConfusionPart2} 
			\begin{center}
				\resizebox{\textwidth}{!}{\begin{tabular}{|l|l|l|||l|l||l|l||l|l|}
						\hline
						Relation                     & \shortstack{\# Correct \\ Predicted} & Rate  &\shortstack{Wrong\\ Relation 1} & \shortstack{\# False\\ Predicted \\ for \\Relation 1}&\shortstack{Wrong \\Relation 2} & \shortstack{\# False \\Predicted\\ for \\Relation 2 }& \shortstack{Wrong \\Relation 3 }            & \shortstack{\# False \\Predicted \\for\\ Relation 3} \\ \hline
						atlocation                & 172                  & 0.67  & antonym      & 20                             & Usedfor       & 17                        &                           &                                \\ \hline
						desire                    & 36                   & 0.593 & isa           & 6                              & Capableof     & 6                         & Usedfor                   & 5                              \\ \hline
						hasprerequest             & 23                   & 0.547 & sysnonymy     & 4                              & antonym      & 3                         & atlocation                & 2                              \\ \hline
						causesdesire                & 17                   & 0.548 & usedfor       & 7                              &               &                           &                           &                                \\ \hline
						isa                       & 147                  & 0.492 & atlocation    & 26                             & antonym      & 22                        & instanceof                & 22                             \\ \hline
						antonym                  & 68                   & 0.492 & isa           & 17                             & atlocation    & 9                         &                           &                                \\ \hline
						instandof                 & 46                   & 0.479 & isa           & 27                             & atlocation    & 8                         &                           &                                \\ \hline
 						usedfor                   & 47                   & 0.475 & atlocation    & 26                             & isa           & 18                        &                           &                                \\ \hline
 						hasa                      & 24                   & 0.393 & antonym      & 11                             & usedfor       & 6                         &                           &                                \\ \hline
						hassubevent               & 12                   & 0.378 & causes        & 5                              & antonym        & 4                         &                           &                                \\ \hline
						partof                    & 16                   & 0.374 & synonym     & 12                             & antonym        & 3                         & hasproperty               & 3                              \\ \hline
						haspropertry              & 12                   & 0.375 & isa           & 8                              &               &                           &                           &                                \\ \hline
						sysnonym                  & 54                   & 0.312 & isa           & 31                             & hasproperty   & 17                        & atlocation                & 12                             \\ \hline
						derivedfrom               & 20                   & 0.307 & isa           & 10                             & sysnonym      & 8                         &\shortstack{etymologically-\\derivedfrom} & 8                              \\ \hline
 						\shortstack{etymologically-\\derivedfrom} & 6                    & 0.3   & derivedfrom   & 6                              &               &                           &                           &                                \\ \hline
 						capableof                 & 13                   & 0.26  & usedfor       & 13                             & isa           & 7                         &                           &                                \\ \hline
 						motivatedbygoal          & 3                    & 0.25  & causes        & 3                              & hassubevent   & 2                         &                           &                                \\ \hline
 						receivsection             & 5                    & 0.238 & atlocation    & 9                              & usedfor       & 3                         &                           &                                \\ \hline
 						createdby                 & 4                    & 0.2   & antonym       & 6                              & isa           & 5                         &                           &                                \\ \hline
 						madeof                    & 3                    & 0.16  & isa           & 7                              & antonym       & 3                         & hsaa                      & 2                              \\ \hline
 						causes                    & 3                    & 0.15  & causesdesire  & 6                              & hassubevent   & 4                         & derivedfrom               & 3                              \\ \hline
					\end{tabular}}
				\end{center}		
			\end{table}
At table \ref{ConfusionPart1} \textit{'Correctly Predicted'} column indicates the proportion of relations are predicted correctly, and \textit{'Correct Prediction Rate'} column indicates the rate of correct predicted. For instance, our model predicts the relation $notisa$ 100 percent correct. 
% Based on table \ref{ConfusionPart1}, amount of some relations are zero that indicates these labels are not in the test set. For other labels whereby their accuracy  was low, we conducted a comprehensive analysis. 
			
Table \ref{ConfusionPart2} shows the relations which are wrongly predicted (\textit{'Wrongly Predicted'} columns).
% shows, some of the labels with \textit{Count of Correct Predicted} and its rates that in table \ref{ConfusionPart1} are shown, plus some more information that is included some incorrect  labels with their count that our model predicted mistakenly. 
Based on the results, the most incorrectly predicted relation is $'isa'$, which accounts for a large proportion of relations of the dataset (around 150 out of 550). In the second place is $'atlocation'$ relation (172 out of 550). The third place is the $'antonym'$ relation. On the other hand, some relations which are correctly unpredicted, can be treated as semantically equivalent to their prediction, where the assignment is dependent on a modelling decision. The same situation occurs for $'etymologicallyderivedfrom'$ and $'derivedfrom'$ relations. \\ Another issue is the low number of certain relations expressed int he dataset.

\section{Conclusion}\label{Conclusion}
In this paper we introduced the task of composite semantic relation classification. The paper proposes a composite semantic relation classification model which combines \textit{commonsense KB lookup}, a \textit{distributional semantic based filter} and the application of a \textit{sequence machine learning model} to address the task. The proposed LSTM model outperformed existing baselines with regard to f1-score, accuracy and recall. Future work will focus on increasing the volume of the training set for under-represented relations.

% SimLex-999\cite{hill2015simlex999} and MEN-3000\cite{bruni}.

%\vfill\eject
\bibliography{llncs}

\begin{thebibliography}{10}
\providecommand{\url}[1]{\texttt{#1}}
\providecommand{\urlprefix}{URL }

\bibitem{barzegar2015dinfra}
Barzegar, S., Sales, J.E., Freitas, A., Handschuh, S., Davis, B.: Dinfra: A one
  stop shop for computing multilingual semantic relatedness. In: Proceedings of
  the 38th International ACM SIGIR Conference on Research and Development in
  Information Retrieval. pp. 1027--1028. ACM (2015)

\bibitem{bordes2013translating}
Bordes, A., Usunier, N., Garcia-Duran, A., Weston, J., Yakhnenko, O.:
  Translating embeddings for modeling multi-relational data. In: Advances in
  Neural Information Processing Systems. pp. 2787--2795 (2013)

\bibitem{freitas2014distributional}
Freitas, A., da~Silva, J.C.P., Curry, E., Buitelaar, P.: A distributional
  semantics approach for selective reasoning on commonsense graph knowledge
  bases. In: Natural Language Processing and Information Systems, pp. 21--32.
  Springer (2014)

\bibitem{garcia2016combining}
Garcia-Duran, A., Bordes, A., Usunier, N., Grandvalet, Y.: Combining two and
  three-way embedding models for link prediction in knowledge bases. Journal of
  Artificial Intelligence Research  55,  715--742 (2016)

\bibitem{hendrickx2009semeval}
Hendrickx, I., Kim, S.N., Kozareva, Z., Nakov, P., {\'O}~S{\'e}aghdha, D.,
  Pad{\'o}, S., Pennacchiotti, M., Romano, L., Szpakowicz, S.: Semeval-2010
  task 8: Multi-way classification of semantic relations between pairs of
  nominals. In: Proceedings of the Workshop on Semantic Evaluations: Recent
  Achievements and Future Directions. pp. 94--99. Association for Computational
  Linguistics (2009)

\bibitem{hochreiter1997long}
Hochreiter, S., Schmidhuber, J.: Long short-term memory. Neural computation
  9(8),  1735--1780 (1997)

\bibitem{jans2012skip}
Jans, B., Bethard, S., Vuli{\'c}, I., Moens, M.F.: Skip n-grams and ranking
  functions for predicting script events. In: Proceedings of the 13th
  Conference of the European Chapter of the Association for Computational
  Linguistics. pp. 336--344. Association for Computational Linguistics (2012)

\bibitem{liu2015dependency}
Liu, Y., Wei, F., Li, S., Ji, H., Zhou, M., Wang, H.: A dependency-based neural
  network for relation classification. arXiv preprint arXiv:1507.04646  (2015)

\bibitem{nguyen2015combining}
Nguyen, T.H., Grishman, R.: Combining neural networks and log-linear models to
  improve relation extraction. arXiv preprint arXiv:1511.05926  (2015)

\bibitem{qin2016empirical}
Qin, P., Xu, W., Guo, J.: An empirical convolutional neural network approach
  for semantic relation classification. Neurocomputing  (2016)

\bibitem{dos2015classifying}
dos Santos, C.N., Xiang, B., Zhou, B.: Classifying relations by ranking with
  convolutional neural networks. In: Proceedings of the 53rd Annual Meeting of
  the Association for Computational Linguistics and the 7th International Joint
  Conference on Natural Language Processing. vol.~1, pp. 626--634 (2015)

\bibitem{socher2013reasoning}
Socher, R., Chen, D., Manning, C.D., Ng, A.: Reasoning with neural tensor
  networks for knowledge base completion. In: Advances in Neural Information
  Processing Systems. pp. 926--934 (2013)

\bibitem{speer2012representing}
Speer, R., Havasi, C.: Representing general relational knowledge in conceptnet
  5. In: LREC. pp. 3679--3686 (2012)

\bibitem{xu2015semantic}
Xu, K., Feng, Y., Huang, S., Zhao, D.: Semantic relation classification via
  convolutional neural networks with simple negative sampling. arXiv preprint
  arXiv:1506.07650  (2015)

\bibitem{xu2016improved}
Xu, Y., Jia, R., Mou, L., Li, G., Chen, Y., Lu, Y., Jin, Z.: Improved relation
  classification by deep recurrent neural networks with data augmentation.
  arXiv preprint arXiv:1601.03651  (2016)

\bibitem{xu2015classifying}
Xu, Y., Mou, L., Li, G., Chen, Y., Peng, H., Jin, Z.: Classifying relations via
  long short term memory networks along shortest dependency paths. In:
  Proceedings of Conference on Empirical Methods in Natural Language Processing
  (to appear) (2015)

\bibitem{zeng2014relation}
Zeng, D., Liu, K., Lai, S., Zhou, G., Zhao, J., et~al.: Relation classification
  via convolutional deep neural network. In: COLING. pp. 2335--2344 (2014)

\end{thebibliography}
\bibliographystyle{splncs03}
\end{document}